\documentclass[runningheads]{llncs}
\usepackage{graphicx}
\usepackage{microtype}
\usepackage{multirow}
\usepackage[utf8]{inputenc}
\usepackage{booktabs}
\usepackage{xspace}
\usepackage{subfig}
\usepackage{amsmath}
\usepackage{relsize}

\usepackage{hyperref}
\usepackage{breakcites}

\newcommand\ie{{\it i.e.\ }}

\begin{document}
\title{LTG-Oslo Hierarchical Multi-task Network:\\
The importance of negation for 
       document-level sentiment in Spanish}
\titlerunning{LTG-Oslo Hierarchical Multi-task Network}
%
\author{Jeremy Barnes\orcidID{0000-0002-8043-8058}}
\authorrunning{J. Barnes}
%
\institute{Language Technology Group \\
University of Oslo, Norway \\
\email{jeremycb@ifi.uio.no}}
\maketitle              
\begin{abstract}
This paper details LTG-Oslo team's participation in the sentiment track of the NEGES 2019 evaluation campaign. We participated in the task with a hierarchical multi-task network, which used shared lower-layers in a deep BiLSTM to predict negation, while the higher layers were dedicated to predicting document-level sentiment. The multi-task component shows promise as a way to incorporate information on negation into deep neural sentiment classifiers, despite the fact that the absolute results on the test set were relatively low for a binary classification task.

\end{abstract}
\section{Introduction}

Sentiment analysis has improved greatly over the last decade, moving from models trained on hand-engineered features \cite{Pang2002,Das2007} to neural models that are trained in an end-to-end fashion \cite{Socher2013b}. The success of these neural architectures is often attributed to their ability to capture compositionality effects \cite{Socher2013b,linzen-etal-2016-assessing}, of which negation is the most common and influential for sentiment analysis \cite{Wiegand2010}. However, recent research has shown that these models are still not able to fully resolve the effect that negation has on sentence-level sentiment \cite{barnes-etal-2019}.

Explicit negation detection has proven useful to create features for lexicon-based sentiment models \cite{councill-etal-2010-great,Cruz2016} and machine-learning approaches to sentiment classification \cite{Lapponi2012}. At the same time, these approaches build upon work on negation detection as its own task \cite{Vin:Sza:Far:08,Morante2012}.

More recent approaches to sentiment, however, have concentrated on learning the effects of negation in an end-to-end fashion. Current state-of-the-art approaches employ neural networks which implicitly learn to resolve negation, by either directly training on sentiment annotated data \cite{Socher2013b,Tai2015a}, or by pre-training the model on a language modeling task \cite{Peters2018,Devlin2019}. State-of-the-art neural methods, however, have not attempted to harness explicit negation detection models and annotated negation datasets to improve results. We hypothesize that multi-task learning (MTL) \cite{Caruana93multitasklearning,Collobert2011a} is an appropriate framework to incorporate negation information into neural models.

In this paper, we propose a multi-task learning approach to explicitly incorporate negation annotated data into a neural sentiment model. We show that this approach improves the final result, although our model performs weakly in absolute terms.

\section{Related Work}
\label{relatedwork}

In this section, we briefly review previous work that is relevant to (\textit{i}) attempts to use negation information in sentiment analysis, (\textit{ii}) research on negation detection as a separate task, and (\textit{iii}) multi-task learning.

\subsection{Negation informed Sentiment Analysis}

Negation is a pervasive linguistic phenomenon which has a direct effect on the sentiment of a text \cite{Wiegand2010}. Take the following example from the SFU ReviewSP-Neg training data, where the negation cue is shown in \textbf{bold} and the scope is \underline{underlined}. 

\begin{example}
\item\label{ex3} El hotel está situado en la puerta de toledo, \textbf{no} \underline{está lejos del centro}. 
\end{example}

The English translation is ``The hotel is located at the \textit{puerta de toledo}, it is not far from the center.'' A sentiment classification model must be able to identify the relevant sentiment words (in this case ``lejos del centro''), negation cues (``no''), and resolve the scope in order to correctly predict that this sentence expresses negative polarity. Intuitively, a sentiment model that has access to negation scope information should perform better than a non-informed version.

The first approaches to detecting negation scope for sentiment used heuristics, such as assuming all tokens between a negation cue and the next punctuation mark are in scope \cite{HuandLiu2004}. However, this simplification does not work well on noisy text, such as tweets, or texts that use more complex syntax, such as those in the political domain.

Later research showed that using machine-learning techniques to detect the scope of negation could improve both lexicon-based \cite{councill-etal-2010-great,Cruz2016} and machine learning  \cite{Lapponi2012} classification of sentiment.

\subsection{Negation detection}

Approaches to negation analysis often decompose the task into two sub-tasks, performing (i) negation cue detection, followed by (ii) scope detection.

Much work was done within the biomedical domain \cite{Mor:Lie:Dae:08,Mor:Dae:09,VelOvrRea12} due largely to the availability of the BioScope corpus \cite{Vin:Sza:Far:08}, which is annotated for negation cues and scopes. The *SEM shared task \cite{Morante2012} instead focused on detection of negation cues and scopes in a corpus of sentences taken from the works of Aurthur Conan Doyle.

Traditional approaches to the task of negation detection have typically employed a wide range of hand-crafted features describing a number of both lexical, morphosyntactic and even semantic properties of the text \cite{ReaVelOvr12b,Pac:Ben:Rea:2014,Lapponi2012,Whi:2012,EngVelOvr17}. \\More recently, research has moved towards using neural models such as CNNs \cite{Qia:Li:Zhu:2016}, feed-forward networks, or LSTMs \cite{Fancellu2016}, finding that these architectures often outperform the previous methods, while requiring less hand-crafting of features.

\subsection{Multi-task learning}

Multi-task learning (MTL) is an approach to machine learning where a single model is trained simultaneously on two tasks. By restricting the search space of possible representations to those that are predictive for both tasks, we attempt to give the model a useful inductive bias \cite{Caruana93multitasklearning}.

\textit{Hard parameter sharing} \cite{Caruana93multitasklearning}, which assumes that all layers are shared between tasks except for the final predictive layer, is the simplest way to implement a multi-task model. When the main task and auxiliary task are closely related, this approach has been shown to be an effective way to improve model performance \cite{Collobert2011a,peng-dredze-2017-multi,Alonso2017,Augenstein2018}. On the other hand, \cite{Sogaard2016} find that it is better to make predictions for low-level auxiliary tasks at lower layers of a multi-layer MTL setup. They also suggest that under the hard-parameter framework auxiliary tasks need to be sufficiently similar to the main task for MTL to improve over the single-task baseline.

In this work, we implement a multi-task learning where the lower layers of a deep neural network are shared for the main and auxiliary tasks (in our case sentiment classification and negation detection, respectively), while higher layers are allowed to adapt to the main task.

\section{Model}

\begin{figure*}[h!]
\centering
\includegraphics[width=\textwidth]{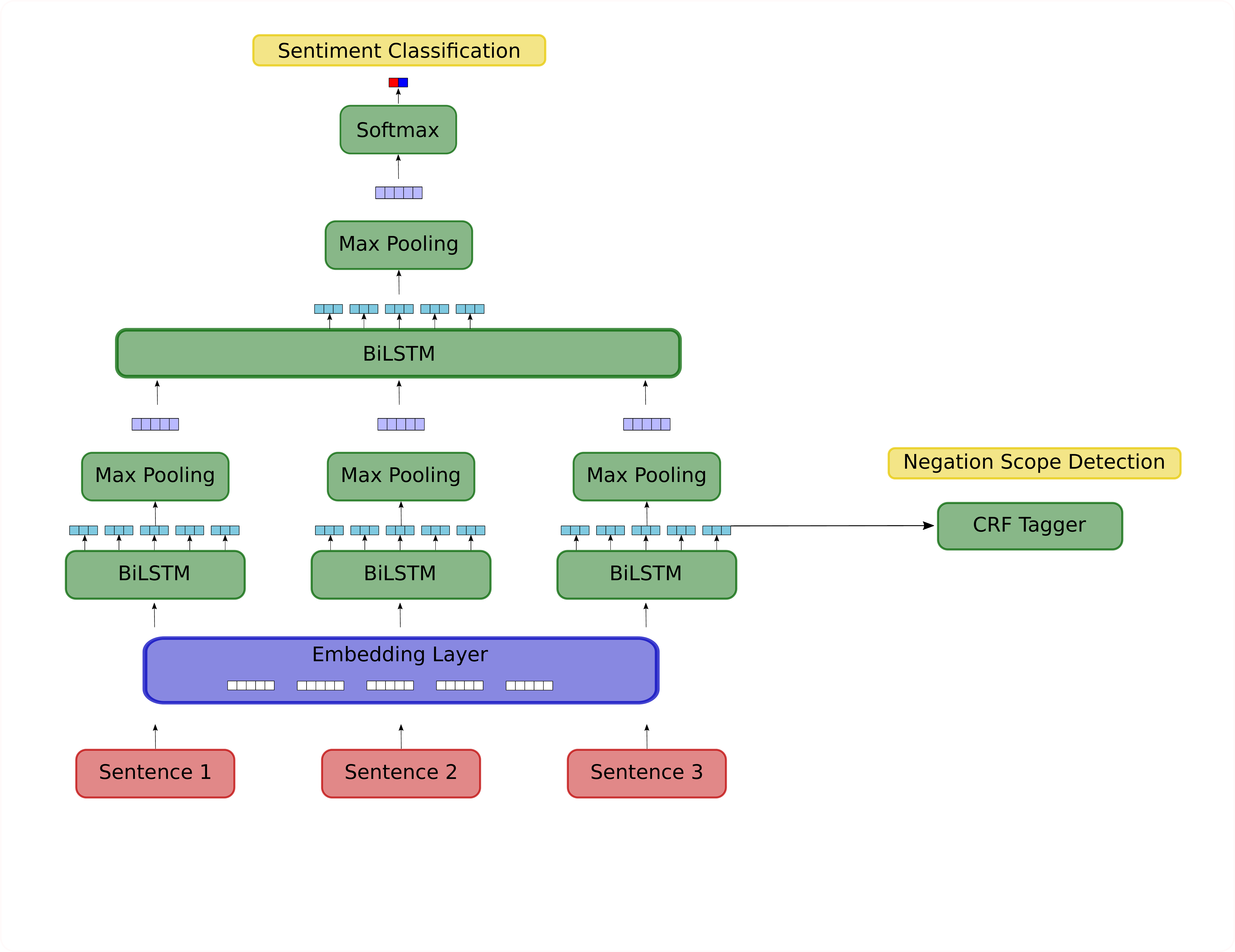}
\caption{Hierarchical multi-task model. The lower BiLSTM is used both to perform sequence-tagging of negation, as well as creating sentence-level features. These features are then aggregated using a second BiLSTM layer and used for predicting the sentiment at document-level.}
\label{fig:model}
\end{figure*}

We propose a \emph{hierarchical multi-task model} (see Figure \ref{fig:model}) which relies on a BiLSTM to create a representation for each sentence in a document, and a second BiLSTM to aggregate these sentence representations into a full document representation. In this section, we first describe the negation submodel, then the sentiment submodel, and finally the multi-task model.

\subsection{Negation Model}

\begin{figure*}[h!]
\centering
\includegraphics[width=\textwidth]{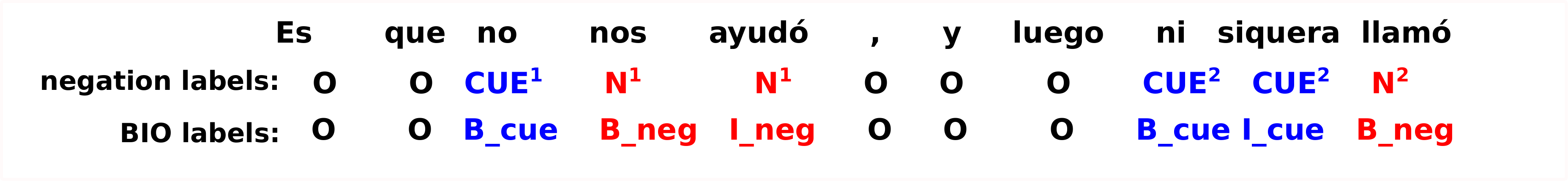}
\caption{An example of the negation which has been converted to BIO labels. Although the example here shows two negation structures where the cue is at the beginning of the scope, there are also examples where the scope begins before the cue or is discontinuous.}
\label{fig:negexample}
\end{figure*}

In previous work on negation detection, it is common to model negation scope as a two step process, where first the negation cues are identified, and then negation scope is determined. However, we hypothesize that within a multi-task framework, it is more beneficial for a network to learn to both identify cues and resolve scope jointly. Therefore, we model negation as a \emph{sequence labeling task} with BIO tags. In the cases where there are more than one negation scope in a sentence that overlap, we flatten these multiple representations, as shown in Figure \ref{fig:negexample}. The negation model, therefore, attempts to identify all cues and all scopes in a sentence at the same time. Note that scopes can also begin before the negation cue and also be discontinuous. While this is an oversimplification of the full negation scope task, we argue that in order to classify sentiment, it is enough for a model to know which tokens are negated.

The negation model is comprised of an embedding layer which embeds the tokens for each sentence. The embeddings pass to a bidirectional Long Short-Term Memory module (BiLSTM), which creates contextualized representations of each word. A linear chain conditional random field (CRF) uses the output of the BiLSTM layer as features. We use Viterbi decoding and minimize the negative log likelihood of CRF predictions.

\subsection{Sentiment Model}

As mentioned above, the sentiment model uses a hierarchical approach. For each sentence in a document, we first extract features with a BiLSTM. We take the max of the BiLSTM output as a representation for the sentence. This is then passed to a second BiLSTM layer, after which we again take the max. We use a softmax layer to compute the sentiment predictions for each document and minimize the cross entropy loss. As a baseline, we train a single-task sentiment model (STL) on the available sentiment data.

\subsection{Multi-task Model}

For the hierarchical multi-task model (MTL), we train both tasks simultaneously by sequentially training the negation classification model for one full epoch and then training the sentiment model. We use Adam as an optimizer, and a dropout layer (0.3) after the embedding layer to regularize the model, as this is common for both the main and auxiliary tasks.

\section{Experimental Setup}
\label{expsetup}

Given that neural models are sensitive to random initialization, we perform five runs for each model on the development data with different random seeds and report both mean accuracy and standard deviation across the five runs. As the final submission required a single prediction for each document, we take a majority vote of the five learned classifiers in order to provide an ensemble prediction.

Besides the proposed STL and MTL models, we also compare with a baseline (BOW) which uses an L2 regularized logistic regression classifier trained on a bag-of-words representation of the documents. We choose the optimal $C$ parameter on the development data.

\subsection{Dataset}

The SFU ReviewSP-NEG dataset \cite{Jimenez-Zafra2018} provided in the shared task contains 400 Spanish-language reviews from eight domains (books, cars, cellphones, computers, hotels, movies, music, and washing machines) which also contain annotations for negation cues, negation scope, and relevance of the negation to sentiment. The participants were provided with the train and dev splits, while the test split was kept from participants until after the final results were posted. Table \ref{table:stats} shows the statistics of the dataset.

Previous work \cite{JIMENEZZAFRA2018240} reported Macro F1 score of 75.89 when using a Bayesian logistic regression classifier trained with bag-of-words features plus negation features that indicate that negation changes the polarity of the negated phrase. However, these results are not comparable to those obtained in the shared task, as the authors evaluated their model using 10-fold cross-validation and not on the test set provided by the organizers. Additionally, they had access to negation information in the test set, which participants in the shared task do not.

\begin{table*}
\centering
\begin{tabular}{llll}
\toprule
Task & Train & Dev & Test\\
\cmidrule(lr){1-1}\cmidrule(lr){2-2}\cmidrule(lr){3-3}\cmidrule(lr){4-4}
Document-level Sentiment & 264 & 56 & 80 \\
Negative Structures & 2,733 & 645 & 949 \\
\bottomrule
\end{tabular}
\caption{Statistics of the document-level sentiment (number of documents) and negation (number of negation structures) data provided by the organizers of the shared task.}
\label{table:stats}
\end{table*}

\subsection{Model performance}

As we only had access to the gold labels on the development set, we report the mean and average accuracy of all three models (BOW, STL, MTL) in Table \ref{table:results}. Additionally, we show the official accuracy score of the MTL model on the test set\footnote{Note that we do not have access to the gold sentiment or negation labels on the test set, so we cannot perform multiple runs, but must rely on the organizers evaluation.}. BOW and STL achieve the same performance, with 71.4 accuracy on the dev set. MTL improves 1.1 percentage points over the other two models on the dev set, and reaches 66.2 accuracy on the test set. In absolute terms, the performance of all models is weak for a binary document-level classification task. This is likely due to the small number of training examples available, as well as the number of domains, which has been shown to be more problematic for machine-learning approaches than lexicon-based approaches \cite{Taboada2011}. 

\begin{table}
\centering
\begin{tabular}{lll}
\toprule
Model  & Dev & Test\\
\cmidrule(lr){1-1}\cmidrule(lr){2-2}\cmidrule(lr){3-3}
BOW & 71.4 & -- \\
STL & 71.4 (5.2) & -- \\
MTL & 72.5 (1.8) & 66.2 \\
\bottomrule
\end{tabular}
\caption{Accuracy of the models on the development and test data. Neural models also report mean accuracy and standard deviation on the development data over five runs with different random seeds.}
\label{table:results}
\end{table}

\subsection{Error Analysis}

\begin{figure*}[h!]
\centering
\includegraphics[width=0.9\textwidth]{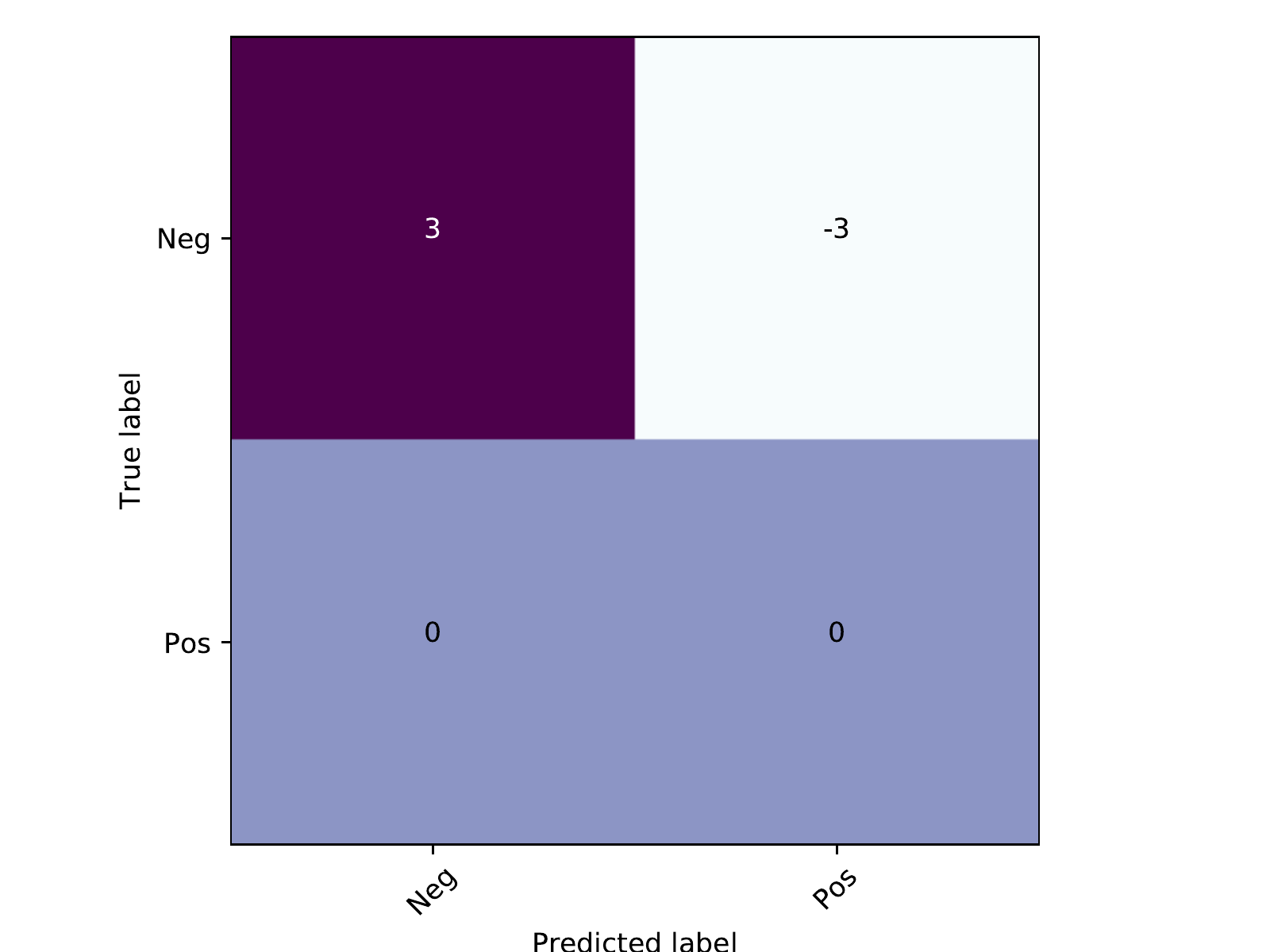}
\caption{A relative confusion matrix, where positive numbers (dark purple) indicate that the MTL model made more predictions in that square that the STL model and negative numbers (white) indicate fewer predictions.}
\label{fig:cm}
\end{figure*}

Given that the classification task is performed at document-level, it is often difficult to determine what exactly was the cause of a change in prediction from one model to another. Instead, Figure \ref{fig:cm} shows a relative confusion matrix of the development results, where positive numbers (dark purple) indicate that the MTL model made more predictions in that square than the STL model and negative numbers (white) indicate fewer predictions. On the development data, the MTL model tends to help with the negative class, while adding little to the positive class. The number of negation structures per class (shown in Table \ref{table:perclass}) shows that there are more negation structures in documents labeled with negative sentiment in the development set, which seems to corroborate the idea that the MTL model is able to use negation information to improve the results on the negative class.

\begin{table*}
\centering
\begin{tabular}{lll}
\toprule
  & Train & Dev \\
\cmidrule(lr){1-1}\cmidrule(lr){2-2}\cmidrule(lr){3-3}
Positive & 1,421 & 303\\
Negative & 1,312 & 342\\
\bottomrule
\end{tabular}
\caption{Number of negation structures per sentiment class found in the training and development data.}
\label{table:perclass}
\end{table*}

\section{Conclusion and Future Work}
\label{conclusion}

In this paper, we have detailed our participation in the 2019 Neges shared task. Our approach, the hierarchical multi-task negation model, did not give a strong performance in absolute numbers on the test set (66\% accuracy), but does indicate that multi-task learning is an appropriate framework for incorporating negation information into sentiment models, improving from 71.4 to 72.5 accuracy on the development set.

The hierarchical RNN model used in this participation is similar to strong performing approaches at sentence-level. However, it is not clear that it is the most adequate model for document-level classification. Convolutional neural networks \cite{Kim2014} or self-attention networks \cite{Ambartsoumian2018} have shown good performance for text classification and may be better models for document-level sentiment tasks.

Additionally, the small training set size for the sentiment task (271 documents) and number of domains (8)  complicates the use of deep neural architectures. Lexicon-based and linear machine-learning approaches have shown to perform quite well under these circumstances \cite{Taboada2011,Cruz2016}. In the future, it would be interesting to use distant supervision \cite{Tang2014,felbo-etal-2017-using} to augment the sentiment signal, or cross-lingual approaches \cite{Chen2016,Barnes2018b} to improve the results.

In this work we have only explored using a sequence-labeling approach to negation scope. It would be interesting to incorporate state-of-the-art negation scope models \cite{fancellu-etal-2017-detecting} into a multi-task setup. 

Finally, the SFU ReviewSP-NEG dataset has several additional levels of annotation, \ie if a negation structure changes the polarity of the tokens in scope or the final polarity after negation. Future work should explore the use of this information further. 

\section*{Acknowledgements}
 This work has been carried out as part of the SANT project, funded by the Research Council of Norway (grant number 270908).

\bibliography{lit}

\begin{thebibliography}{}

\bibitem[Ambartsoumian and Popowich, 2018]{Ambartsoumian2018}
Ambartsoumian, A. and Popowich, F. (2018).
\newblock Self-attention: A better building block for sentiment analysis neural
  network classifiers.
\newblock In {\em Proceedings of the 9th Workshop on Computational Approaches
  to Subjectivity, Sentiment and Social Media Analysis}, pages 130--139.
  Association for Computational Linguistics.

\bibitem[Augenstein et~al., 2018]{Augenstein2018}
Augenstein, I., Ruder, S., and S{\o}gaard, A. (2018).
\newblock Multi-task learning of pairwise sequence classification tasks over
  disparate label spaces.
\newblock In {\em Proceedings of the 2018 Conference of the North American
  Chapter of the Association for Computational Linguistics: Human Language
  Technologies, Volume 1 (Long Papers)}, pages 1896--1906. Association for
  Computational Linguistics.

\bibitem[Barnes et~al., 2018]{Barnes2018b}
Barnes, J., Klinger, R., and Schulte~im Walde, S. (2018).
\newblock Bilingual sentiment embeddings: Joint projection of sentiment across
  languages.
\newblock In {\em Proceedings of the 56th Annual Meeting of the Association for
  Computational Linguistics (Volume 1: Long Papers)}, pages 2483--2493.
  Association for Computational Linguistics.

\bibitem[Barnes et~al., 2019]{barnes-etal-2019}
Barnes, J., Øvrelid, L., and Velldal, E. (2019).
\newblock Sentiment analysis is not solved!: Assessing and probing sentiment
  classifiers.
\newblock In {\em Proceedings of the 2018 {EMNLP} Workshop {B}lackbox{NLP}:
  Analyzing and Interpreting Neural Networks for {NLP}}, page to appear,
  Florence, Italy. Association for Computational Linguistics.

\bibitem[Caruana, 1993]{Caruana93multitasklearning}
Caruana, R. (1993).
\newblock Multitask learning: A knowledge-based source of inductive bias.
\newblock In {\em Proceedings of the Tenth International Conference on Machine
  Learning}, pages 41--48. Morgan Kaufmann.

\bibitem[Chen et~al., 2016]{Chen2016}
Chen, X., Athiwaratkun, B., Sun, Y., Weinberger, K.~Q., and Cardie, C. (2016).
\newblock Adversarial deep averaging networks for cross-lingual sentiment
  classification.
\newblock {\em CoRR}, abs/1606.01614.

\bibitem[Collobert et~al., 2011]{Collobert2011a}
Collobert, R., Weston, J., Bottou, L., Karlen, M., Kavukcuoglu, K., and Kuksa,
  P. (2011).
\newblock {Natural language processing (almost) from scratch}.
\newblock {\em Journal of Machine Learning Research}, 12:2493--2537.

\bibitem[Councill et~al., 2010]{councill-etal-2010-great}
Councill, I., McDonald, R., and Velikovich, L. (2010).
\newblock What's great and what's not: learning to classify the scope of
  negation for improved sentiment analysis.
\newblock In {\em Proceedings of the Workshop on Negation and Speculation in
  Natural Language Processing}, pages 51--59, Uppsala, Sweden. University of
  Antwerp.

\bibitem[Cruz et~al., 2016]{Cruz2016}
Cruz, N.~P., Taboada, M., and Mitkov, R. (2016).
\newblock A machine-learning approach to negation and speculation detection for
  sentiment analysis.
\newblock {\em Journal of the Association for Information Science and
  Technology}, 67(9):2118--2136.

\bibitem[Das and Chen, 2007]{Das2007}
Das, S.~R. and Chen, M.~Y. (2007).
\newblock Yahoo! for amazon: Sentiment extraction from small talk on the web.
\newblock {\em Management Science}, 53(9):1375--1388.

\bibitem[Devlin et~al., 2019]{Devlin2019}
Devlin, J., Chang, M.-W., Lee, K., and Toutanova, K. (2019).
\newblock {BERT}: Pre-training of deep bidirectional transformers for language
  understanding.
\newblock In {\em Proceedings of the 2019 Conference of the North {A}merican
  Chapter of the Association for Computational Linguistics: Human Language
  Technologies, Volume 1 (Long and Short Papers)}, pages 4171--4186,
  Minneapolis, Minnesota. Association for Computational Linguistics.

\bibitem[Enger et~al., 2017]{EngVelOvr17}
Enger, M., Velldal, E., and {\O}vrelid, L. (2017).
\newblock An open-source tool for negation detection: a maximum-margin
  approach.
\newblock In {\em Proceedings of the EACL workshop on {C}omputational
  {S}emantics {B}eyond {E}vents and {R}oles ({SemBEaR})}, pages 64--69,
  Valencia, Spain.

\bibitem[Fancellu et~al., 2016]{Fancellu2016}
Fancellu, F., Lopez, A., and Webber, B. (2016).
\newblock Neural networks for negation scope detection.
\newblock In {\em Proceedings of the 54th Annual Meeting of the Association for
  Computational Linguistics (Volume 1: Long Papers)}, pages 495--504, Berlin,
  Germany. Association for Computational Linguistics.

\bibitem[Fancellu et~al., 2017]{fancellu-etal-2017-detecting}
Fancellu, F., Lopez, A., Webber, B., and He, H. (2017).
\newblock Detecting negation scope is easy, except when it isn{'}t.
\newblock In {\em Proceedings of the 15th Conference of the {E}uropean Chapter
  of the Association for Computational Linguistics: Volume 2, Short Papers},
  pages 58--63, Valencia, Spain. Association for Computational Linguistics.

\bibitem[Felbo et~al., 2017]{felbo-etal-2017-using}
Felbo, B., Mislove, A., S{\o}gaard, A., Rahwan, I., and Lehmann, S. (2017).
\newblock Using millions of emoji occurrences to learn any-domain
  representations for detecting sentiment, emotion and sarcasm.
\newblock In {\em Proceedings of the 2017 Conference on Empirical Methods in
  Natural Language Processing}, pages 1615--1625, Copenhagen, Denmark.
  Association for Computational Linguistics.

\bibitem[Hu and Liu, 2004]{HuandLiu2004}
Hu, M. and Liu, B. (2004).
\newblock Mining opinion features in customer reviews.
\newblock In {\em Proceedings of the 10th ACM SIGKDD International Conference
  on Knowledge Discovery and Data Mining (KDD 2004)}, pages 168--177.

\bibitem[Jim{\'e}nez-Zafra et~al., 2018a]{JIMENEZZAFRA2018240}
Jim{\'e}nez-Zafra, S.~M., Mart{\'i}n-Valdivia, M.~T., Molina-Gonz{\'a}lez,
  M.~D., and Ure{\~{n}}a-L{\'o}pez, L.~A. (2018a).
\newblock {Relevance of the SFU ReviewSP-NEG corpus annotated with the scope of
  negation for supervised polarity classification in Spanish}.
\newblock {\em Information Processing \& Management}, 54(2):240 -- 251.

\bibitem[Jim{\'e}nez-Zafra et~al., 2018b]{Jimenez-Zafra2018}
Jim{\'e}nez-Zafra, S.~M., Taul{\'e}, M., Mart{\'i}n-Valdivia, M.~T.,
  Ure{\~{n}}a-L{\'o}pez, L.~A., and Mart{\'i}, M.~A. (2018b).
\newblock {SFU ReviewSP-NEG: a Spanish corpus annotated with negation for
  sentiment analysis. A typology of negation patterns}.
\newblock {\em Language Resources and Evaluation}, 52(2):533--569.

\bibitem[Kim, 2014]{Kim2014}
Kim, Y. (2014).
\newblock Convolutional neural networks for sentence classification.
\newblock In {\em Proceedings of the 2014 Conference on Empirical Methods in
  Natural Language Processing (EMNLP)}.

\bibitem[Lapponi et~al., 2012]{Lapponi2012}
Lapponi, E., Read, J., and {\O}vrelid, L. (2012).
\newblock Representing and resolving negation for sentiment analysis.
\newblock In {\em Proceedings of the 2012 IEEE 12th International Conference on
  Data Mining Workshops}, ICDMW '12, pages 687--692, Washington, DC, USA. IEEE
  Computer Society.

\bibitem[Linzen et~al., 2016]{linzen-etal-2016-assessing}
Linzen, T., Dupoux, E., and Goldberg, Y. (2016).
\newblock Assessing the ability of {LSTM}s to learn syntax-sensitive
  dependencies.
\newblock {\em Transactions of the Association for Computational Linguistics},
  4:521--535.

\bibitem[Mart{\'i}nez~Alonso and Plank, 2017]{Alonso2017}
Mart{\'i}nez~Alonso, H. and Plank, B. (2017).
\newblock When is multitask learning effective? semantic sequence prediction
  under varying data conditions.
\newblock In {\em Proceedings of the 15th Conference of the European Chapter of
  the Association for Computational Linguistics: Volume 1, Long Papers}, pages
  44--53. Association for Computational Linguistics.

\bibitem[Morante and Blanco, 2012]{Morante2012}
Morante, R. and Blanco, E. (2012).
\newblock *{SEM} 2012 shared task: Resolving the scope and focus of negation.
\newblock In {\em *{SEM} 2012: The First Joint Conference on Lexical and
  Computational Semantics {--} Volume 1: Proceedings of the main conference and
  the shared task, and Volume 2: Proceedings of the Sixth International
  Workshop on Semantic Evaluation ({S}em{E}val 2012)}, pages 265--274,
  Montr{\'e}al, Canada. Association for Computational Linguistics.

\bibitem[Morante and Daelemans, 2009]{Mor:Dae:09}
Morante, R. and Daelemans, W. (2009).
\newblock A metalearning approach to processing the scope of negation.
\newblock In {\em Proceedings of the Thirteenth Conference on Computational
  Natural Language Learning ({CoNLL})}.

\bibitem[Morante et~al., 2008]{Mor:Lie:Dae:08}
Morante, R., Liekens, A., and Daelemans, W. (2008).
\newblock Learning the scope of negation in biomedical texts.
\newblock In {\em Proceedings of the Conference on Empirical Methods in Natural
  Language Processing ({EMNLP})}.

\bibitem[Packard et~al., 2014]{Pac:Ben:Rea:2014}
Packard, W., Bender, E.~M., Read, J., Oepen, S., and Dridan, R. (2014).
\newblock Simple negation scope resolution through deep parsing: A semantic
  solution to a semantic problem.
\newblock In {\em Proceedings of the 52nd Annual Meeting of the Association for
  Computational Linguistics}.

\bibitem[Pang et~al., 2002]{Pang2002}
Pang, B., Lee, L., and Vaithyanathan, S. (2002).
\newblock Thumbs up? sentiment classification using machine learning
  techniques.
\newblock In {\em Proceedings of the ACL-02 Conference on Empirical methods in
  natural language processing-Volume 10}, pages 79--86. Association for
  Computational Linguistics.

\bibitem[Peng and Dredze, 2017]{peng-dredze-2017-multi}
Peng, N. and Dredze, M. (2017).
\newblock Multi-task domain adaptation for sequence tagging.
\newblock In {\em Proceedings of the 2nd Workshop on Representation Learning
  for {NLP}}, pages 91--100, Vancouver, Canada. Association for Computational
  Linguistics.

\bibitem[Peters et~al., 2018]{Peters2018}
Peters, M., Neumann, M., Iyyer, M., Gardner, M., Clark, C., Lee, K., and
  Zettlemoyer, L. (2018).
\newblock Deep contextualized word representations.
\newblock In {\em Proceedings of the 2018 Conference of the North American
  Chapter of the Association for Computational Linguistics: Human Language
  Technologies, Volume 1 (Long Papers)}, pages 2227--2237. Association for
  Computational Linguistics.

\bibitem[Qian et~al., 2016]{Qia:Li:Zhu:2016}
Qian, Z., Li, P., Zhu, Q., Zhou, G., Luo, Z., and Luo, W. (2016).
\newblock Speculation and negation scope detection via convolutional neural
  networks.
\newblock In {\em The 2016 Conference on Empirical Methods in Natural Language
  Processing}.

\bibitem[Read et~al., 2012]{ReaVelOvr12b}
Read, J., Velldal, E., {\O}vrelid, L., and Oepen, S. (2012).
\newblock {UiO}1: Constituent-based discriminative ranking for negation
  resolution.
\newblock In {\em Proceedings of the First Joint Conference on Lexical and
  Computational Semantics ({*SEM})}, Montreal.

\bibitem[Socher et~al., 2013]{Socher2013b}
Socher, R., Perelygin, A., Wu, J., Chuang, J., Manning, C., Ng, A., and Potts,
  C. (2013).
\newblock {Recursive deep models for semantic compositionality over a sentiment
  treebank}.
\newblock {\em Proceedings of the EMNLP 2013}, pages 1631--1642.

\bibitem[S{\o}gaard and Goldberg, 2016]{Sogaard2016}
S{\o}gaard, A. and Goldberg, Y. (2016).
\newblock Deep multi-task learning with low level tasks supervised at lower
  layers.
\newblock In {\em Proceedings of the 54th Annual Meeting of the Association for
  Computational Linguistics (Volume 2: Short Papers)}, pages 231--235.
  Association for Computational Linguistics.

\bibitem[Taboada et~al., 2011]{Taboada2011}
Taboada, M., Brooke, J., Tofiloski, M., Voll, K., and Stede, M. (2011).
\newblock Lexicon-based methods for sentiment analysis.
\newblock {\em Computational Linguistics}, 37(2):267--307.

\bibitem[Tai et~al., 2015]{Tai2015a}
Tai, K.~S., Socher, R., and Manning, C.~D. (2015).
\newblock {Improved semantic representations From tree-structured long
  short-term memory networks}.
\newblock {\em Association for Computational Linguistics 2015 Conference},
  pages 1556--1566.

\bibitem[Tang et~al., 2014]{Tang2014}
Tang, D., Wei, F., Yang, N., Zhou, M., Liu, T., and Qin, B. (2014).
\newblock Learning sentiment-specific word embedding for twitter sentiment
  classification.
\newblock In {\em Proceedings of the 52nd Annual Meeting of the Association for
  Computational Linguistics (Volume 1: Long Papers)}, pages 1555--1565.

\bibitem[Velldal et~al., 2012]{VelOvrRea12}
Velldal, E., {\O}vrelid, L., Read, J., and Oepen, S. (2012).
\newblock Speculation and negation: Rules, rankers, and the role of syntax.
\newblock {\em Computational Linguistics}, 38(2):369--410.

\bibitem[Vincze et~al., 2008]{Vin:Sza:Far:08}
Vincze, V., Szarvas, G., Farkas, R., Móra, G., and Csirik, J. (2008).
\newblock The bioscope corpus: biomedical texts annotated for uncertainty,
  negation and their scopes.
\newblock {\em BMC bioinformatics}, Suppl 11(Suppl 11).

\bibitem[White, 2012]{Whi:2012}
White, J. (2012).
\newblock {UWashington}: Negation resolution using machine learning methods.
\newblock In {\em Proceedings of the First Joint Conference on Lexical and
  Computational Semantics ({*SEM})}, Montreal.

\bibitem[Wiegand et~al., 2010]{Wiegand2010}
Wiegand, M., Balahur, A., Roth, B., Klakow, D., and Montoyo, A. (2010).
\newblock A survey on the role of negation in sentiment analysis.
\newblock In {\em Proceedings of the Workshop on Negation and Speculation in
  Natural Language Processing}, pages 60--68.

\end{thebibliography}
\end{document}